\documentclass[10pt,twocolumn,letterpaper]{article}

\usepackage[pagenumbers]{cvpr} 

\usepackage{cuted} 

\usepackage{currfile} 

\usepackage{caption} 

\usepackage{multirow}
\usepackage{booktabs}
\usepackage{array}
\usepackage{textcomp}
\usepackage{xcolor}
\usepackage{colortbl}
\usepackage{nicematrix} 
\usepackage{xspace}
\newcommand{\ours}{VERA\xspace}

\definecolor{EviShade}{gray}{0.94}

\usepackage{iftex}
\ifXeTeX
  \usepackage{fontspec}
\fi

\usepackage{tikz}
\usetikzlibrary{arrows.meta,positioning,calc}
\definecolor{ourblue}{HTML}{2A78D6}
\definecolor{ourorange}{HTML}{EB6834}
\definecolor{ouraqua}{HTML}{1BAF7A}
\usepackage{pifont}

\definecolor{cvprblue}{rgb}{0.21,0.49,0.74}
\usepackage[pagebackref,breaklinks,colorlinks,allcolors=cvprblue]{hyperref}

\def\confName{CVPR}
\def\confYear{2027}

\title{Visual Harness Scales General VLMs to Medical Images}

\author{
Shaowen Wan$^{1,*}$ \quad
Hanqi Jiang$^{2,*}$ \quad
Zihao Wu$^{2}$ \quad
Junhao Chen$^{2}$ \quad
Weihang You$^{2}$ \\
Ruiyu Yan$^{3}$ \quad
Bowen Chen$^{1}$ \quad
Tianming Liu$^{2}$ \quad
Lin Zhao$^{1,\dagger}$ \\[4pt]
{\small
$^{1}$New Jersey Institute of Technology \quad
$^{2}$University of Georgia \quad
$^{3}$New York University
} \\[3pt]
{\small
$^{*}$Equal contribution \qquad
$^{\dagger}$\texttt{lin.zhao.1@njit.edu}
}
}

\begin{document}
\maketitle
\begin{abstract}

We show that an appropriate visual harness can substantially improve the performance of general vision-language models (VLMs) across diverse medical imaging tasks. We introduce \ours, a medical visual harness that equips general VLMs with interactive visual tools and systematic diagnostic evidence management. With \ours, general VLMs interactively explore medical images and progressively acquire diagnostic evidence, which is preserved in a lossless memory for subsequent retrieval and use. After submitting an initial answer, the VLM receives an assessment of potential omissions and the reliability of identified findings. This assessment guides further image inspection and measurement, followed by answer revision in an iterative agentic loop until sufficient evidence is obtained. Across 11 tasks and four imaging modalities, \ours improves performance on all tasks with Qwen3.8-27B, with notable gains in spleen measurement (63.3\% to 98.3\%), renal search (57.5\% to 71.2\%), and pulmonary nodule detection (0.0\% to 28.2\%). High-capability API-based VLMs achieve even higher  performance with \ours. For example, GPT-6.1 Sol improves pulmonary nodule detection from 10.8\% to 69.2\% with \ours, demonstrating the complementary benefits of stronger model capabilities and an effective medical visual harness. Our results highlight the potential of visual harnesses as a new paradigm for scaling general VLMs to medical imaging tasks.

\end{abstract}

    
\section{Introduction}
\label{sec:intro}

\begin{figure}[t]
    \centering
    \includegraphics[width=\columnwidth]{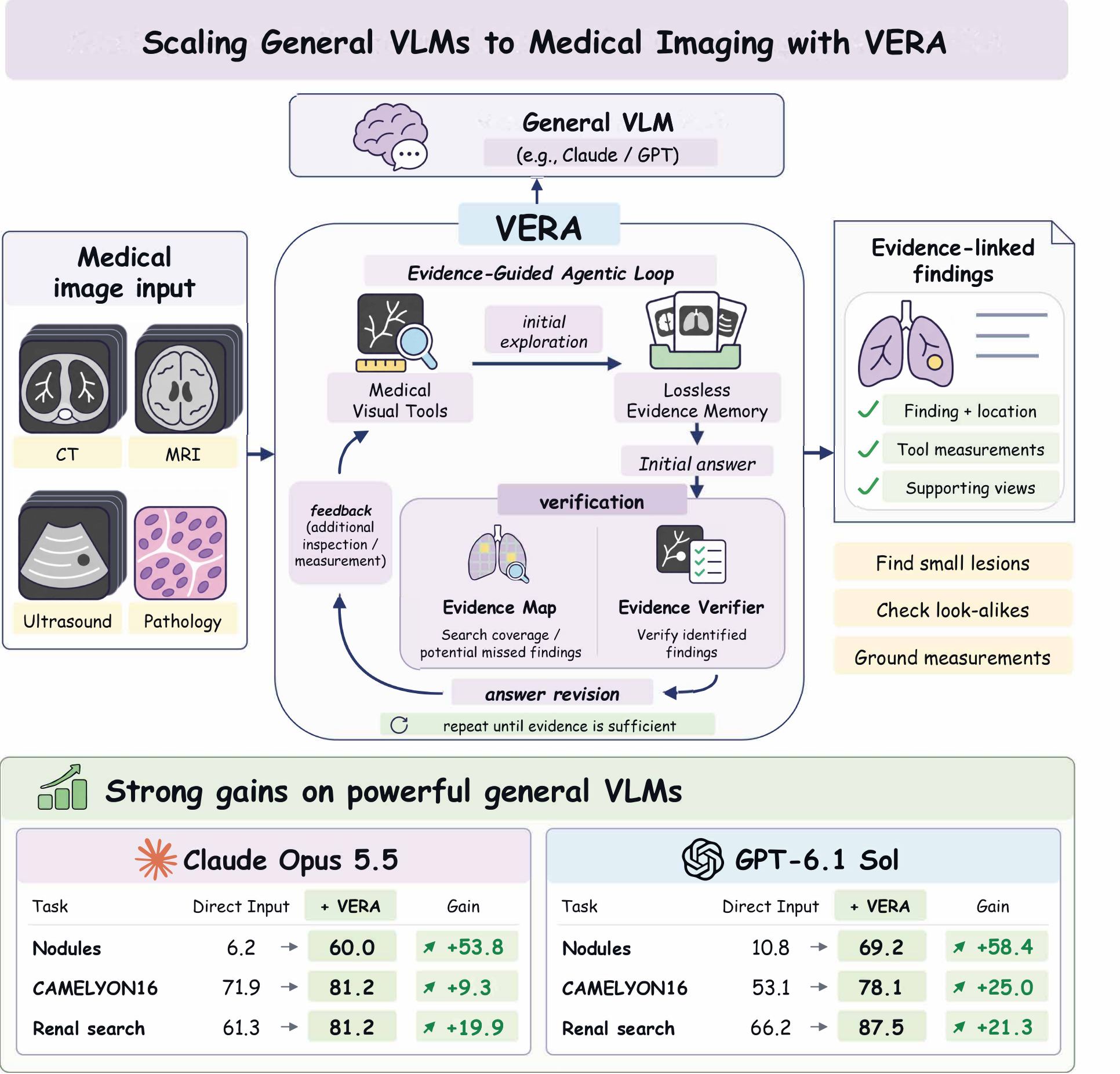}
    \caption{\textbf{Scaling general VLMs to medical imaging with VERA.}
    VERA provides a medical visual harness for image exploration and diagnostic evidence management, delivering substantial performance gains across diverse VLM backbones and medical imaging tasks.}
    \label{fig:teaser}
\end{figure}

General vision-language models (VLMs) have demonstrated strong visual understanding, grounding, and reasoning capabilities across a wide range of tasks~\cite{bai2025qwen3vl,wang2025internvl35,comanici2025gemini25,gemma2026gemma4}. However, these capabilities do not consistently translate into strong performance on medical imaging tasks, a gap commonly attributed to the domain shift between natural and medical images~\cite{li2023llavamed,lingshu}.
To bridge this gap, a major line of research has focused on developing medical VLMs through pretraining and finetuning on domain-specific data~\cite{li2023llavamed,moor2023medflamingo,medgemma,sellergren2026medgemma15,lingshu,hulumed,clinfusion}. Medical AI agents offer a complementary approach, using specialized tools, interactive image exploration, or multi-agent collaboration to enhance the capabilities of VLMs in medical imaging~\cite{li2024mmedagent,medrax,wang2025medagentpro,radagent,wang20263dmedagent,qian2026actor,mdagents,xia2025mmedagentrl}.
 These efforts largely emphasize domain specialization, either in the VLMs themselves or in the systems built around them. Could general VLMs already possess substantial capabilities for medical imaging that remain untapped?

We posit that an appropriate visual harness can substantially improve the performance of general VLMs on medical images by making more effective use of their existing capabilities. A visual harness provides an inference-time interaction mechanism between the model and visual data, organizing how the model accesses, inspects, retrieves, and utilizes visual information throughout the reasoning process. Recent work such as VISTA~\cite{vista} has demonstrated the potential of this paradigm by equipping VLMs with lossless visual memory and tools for active inspection and retrieval, improving their performance in interactive visual environments. However, in medical imaging, visual observations do not necessarily constitute sufficient diagnostic evidence. A medical visual harness should extend beyond managing visual observations to organizing the acquisition and verification of diagnostic evidence.

Effective management of diagnostic evidence involves three complementary aspects: evidence acquisition, preservation, and assessment. First, radiologists acquire evidence through iterative visual inspection, adjusting image magnification, reviewing contiguous slices, examining multiplanar views, and obtaining quantitative measurements~\cite{vanderGijp2017visualsearch,kelahan2019radiologistsgaze}. A medical visual harness should equip VLMs with similar capabilities. Second, diagnostic evidence is accumulated progressively, with previously acquired evidence often revisited for comparison and subsequent assessment. Therefore, all acquired evidence should be preserved losslessly and remain accessible throughout the reading process. Third, lesions may remain undetected in unexamined regions or in regions inspected at insufficient resolution~\cite{brunye2019eyetracking,degnan2019perceptual}. The reliability of identified findings, in turn, is influenced by the level of detail in visual inspection, the viewing perspectives, and the availability of quantitative measurements. A medical visual harness should assess potential omissions and finding reliability to inform further inspection or establish the sufficiency of available evidence.

To address these requirements, we introduce Visual Evidence Reasoning Architecture (VERA), a medical visual harness for general VLMs. VERA consists of five components: Medical Visual Tools, Lossless Evidence Memory, Evidence Map, Evidence Verifier, and an Evidence-Guided Agentic Loop. The VLM uses visual tools to explore medical images and obtain quantitative measurements, while the memory preserves acquired evidence and the Evidence Map tracks the inspected regions and potential omissions. After submitting an initial answer, the VLM receives feedback from the Evidence Map on remaining search regions and from the Evidence Verifier on the reliability of identified findings. This feedback guides further inspection, measurement, and answer revision through the Agentic Loop.

We evaluate \ours on 11 tasks spanning four medical imaging modalities. With Qwen3.8-27B, \ours improves over direct image input on all 11 tasks and achieves best or tied-best performance among agent baselines on 10 tasks. Spleen measurement accuracy rises from 63.3\% to 98.3\%, and pulmonary nodule detection sensitivity from 0.0\% to 28.2\%. Across five additional backbones, \ours improves all four task categories, with stronger models achieving higher absolute performance on multiple tasks. Pulmonary nodule detection sensitivity reaches 60.0\% with Claude Opus 5.5 and 69.2\% with GPT-6.1 Sol, demonstrating the complementary benefits of stronger VLMs and an effective visual harnesses. Overall, our findings highlight the potential of visual harnesses for scaling general VLMs to medical imaging tasks.

\begin{figure*}[t]
\centering
\includegraphics[width=\textwidth]{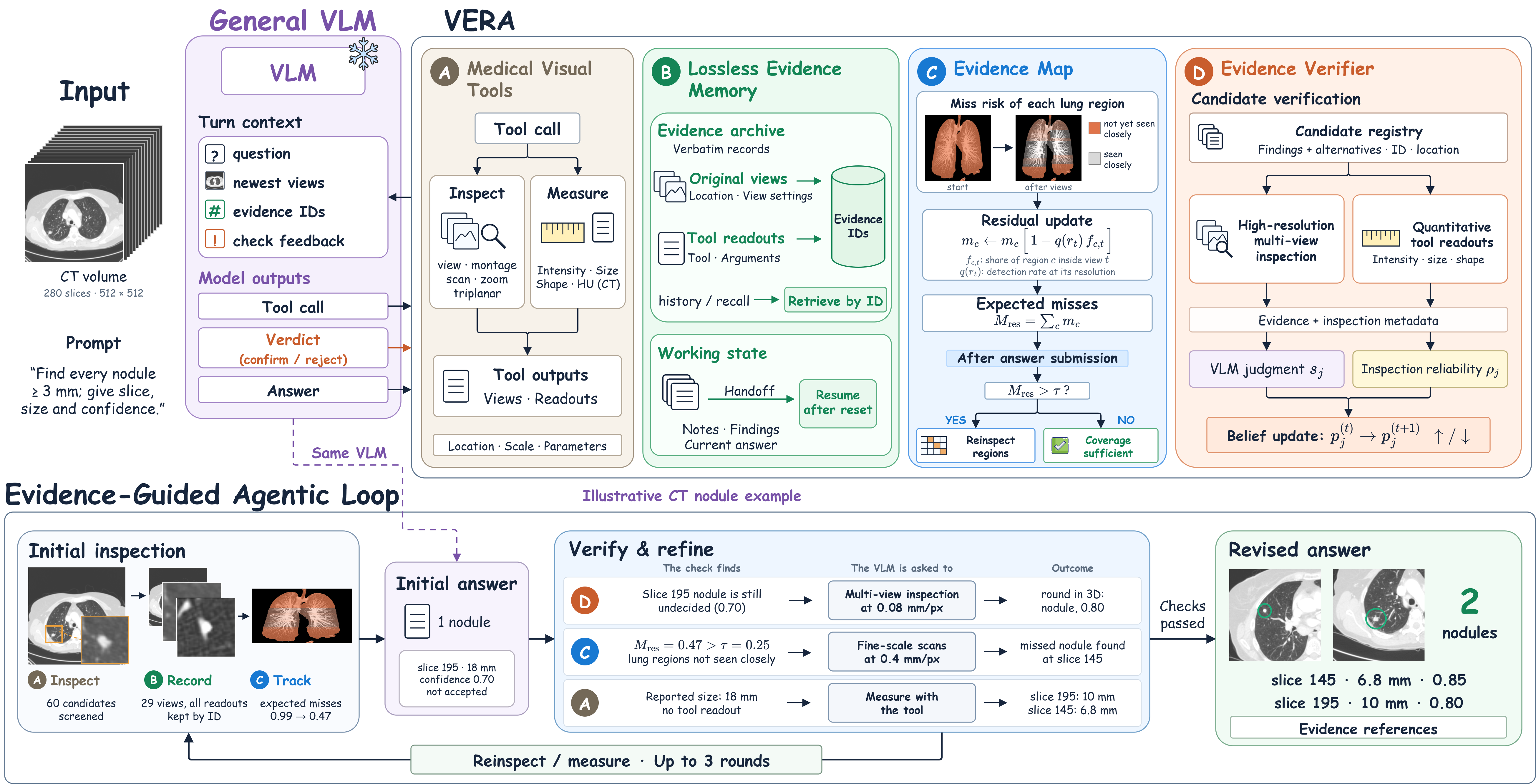}
\caption{\textbf{Overview of \ours.}
\textbf{Top:} a frozen general VLM works with four modules.
(A)~Medical Visual Tools inspect and measure the image.
(B)~Lossless Evidence Memory stores views and tool readouts verbatim for later recall.
(C)~Evidence Map tracks the residual estimate $m_c$ of undetected findings per cell and requests re-inspection when the total $M_{\mathrm{res}}$ exceeds the tolerance $\tau$.
(D)~Evidence Verifier updates the belief $p_j$ of each candidate finding.
\textbf{Bottom:} the Evidence-Guided Agentic Loop on an illustrative CT nodule case, where each row gives what a check finds, the action requested from the VLM, and the outcome.}
\label{fig:overview}
\end{figure*}


\section{Related Work}
\label{sec:related_work}

\subsection{Medical Vision-Language Models}
\label{sec:related_medical_vlms}

Medical vision-language models (VLMs) have been developed by adapting general multimodal architectures to medical image--text data and tasks. Early systems such as LLaVA-Med and Med-Flamingo use biomedical image--text instruction data or continued pretraining to support medical visual question answering and related tasks~\cite{li2023llavamed,moor2023medflamingo}. Recent models have expanded their capabilities across imaging modalities, spatial dimensions, and clinical tasks. MedGemma provides medical-adapted multimodal models at different scales, with MedGemma 1.5 further extending support to volumetric CT and MRI, whole-slide pathology, and anatomical localization~\cite{medgemma,sellergren2026medgemma15}. Lingshu-32B is a generalist medical VLM designed for unified understanding across more than a dozen medical imaging modalities, including CT, MRI, ultrasound, and histopathology~\cite{lingshu}. Hulu-Med extends medical visual understanding to 2D images, 3D volumes, and video through a unified vision encoder and large-scale multimodal training~\cite{hulumed}. ClinFusion adopts a vision-centric architecture with cascaded spatial-aware feature fusion to support both 2D and native 3D medical image understanding~\cite{clinfusion}. Unlike specialized medical VLMs, VERA enhances general VLMs through interactive visual tools and diagnostic evidence management.

\subsection{Medical AI Agents}
\label{sec:related_medical_agents}

Medical AI agents extend foundation models with tools, task decomposition, or collaboration among multiple agents. MDAgents adapts the collaboration structure of language-model agents to the complexity of medical decision-making tasks~\cite{mdagents}. 3DMedAgent coordinates visual and textual tools to support multi-step analysis of volumetric CT, combining perception modules with an agent that moves between global and local views~\cite{wang20263dmedagent}.  Visual tool-use approaches such as VLM-R$^3$, DeepEyes also let multimodal models select image regions or request additional visual observations during reasoning~\cite{jiang2026vlm,deepeyes}. VERA extends this capability by explicitly assessing the diagnostic evidence acquired during these interactions, including the coverage of image inspection and the reliability of identified findings. This assessment guides further inspection and finding verification within Evidence-Guided Agentic Loop.

\subsection{Visual Harnesses}
\label{sec:related_harnesses}

The capabilities of an LLM depend not only on the model itself but also on the system that supports its interaction with external tools and the environment. An LLM harness provides this support by managing tool execution, memory, and task progression. ReAct~\cite{yao2023react} established a widely adopted reasoning--acting loop in which the model alternates between reasoning, tool execution, and observation. Recently, VISTA~\cite{vista} extends this paradigm to visual tasks by combining active image inspection with lossless visual memory, demonstrating the benefits of organizing visual interaction around frozen VLMs. However, medical imaging tasks require more than visual observation: diagnostic evidence is progressively acquired, preserved, and verified before a diagnostic conclusion is reached. VERA extends the visual harness paradigm from managing visual observations to managing diagnostic evidence throughout the
interpretation process. VERA introduces evidence-guided verification into the agentic loop, using the assessment of acquired evidence to guide subsequent model actions.


\section{Method}
\label{sec:method}

\subsection{Overview}
\label{sec:overview}

We introduce \ours, a medical visual harness that enables general VLMs to acquire, retain, and assess diagnostic evidence for medical imaging tasks. As illustrated in \Cref{fig:overview}, \ours consists of five components: Medical Visual Tools, Lossless Evidence Memory, Evidence Map, Evidence Verifier, and Evidence-Guided Agentic Loop.

The VLM explores the image using Medical Visual Tools for navigation, multi-view inspection, and quantitative measurement (\Cref{sec:visual_tools}), while Lossless Evidence Memory preserves acquired images and measurements (\Cref{sec:evidence_memory}). The Evidence Map tracks inspection coverage (\Cref{sec:evidence_map}). After an initial answer, the Evidence Map identifies regions requiring further inspection, and the Evidence Verifier assesses findings and alternatives (\Cref{sec:evidence_verifier}). Their feedback guides further inspection, measurement, and answer revision in the Evidence-Guided Agentic Loop (\Cref{sec:loop}) until the checks pass or the interaction budget is exhausted.

\subsection{Medical Visual Tools}
\label{sec:visual_tools}

Medical images contain findings across different locations, scales, and views. A single rendered image may not capture small findings or reveal the spatial relationships of anatomical structures. \ours provides visual tools that allow the VLM to navigate medical images, adjust magnification, inspect local regions, and examine structures from multiple views. The \texttt{view} tool supports slice and region inspection with adjustable viewing parameters, while \texttt{montage} and \texttt{scan} enable examination of multiple slices or image regions. For volumetric images, \texttt{triplanar} provides axial, coronal, and sagittal views of the same location.

Visual inspection alone also provides limited access to quantitative properties of medical images. \ours therefore provides measurement tools that extract quantitative information directly from the image data. The \texttt{read\_hu} and \texttt{read\_intensity} tools return intensity statistics for selected regions, while \texttt{measure} and \texttt{read\_shape} provide physical measurements and geometric descriptors of identified structures. Modality-specific tools further support measurements such as cavity dimensions and volume in ultrasound. These tools allow the VLM to combine visual observations with quantitative measurements when assessing medical findings.

\subsection{Lossless Evidence Memory}
\label{sec:evidence_memory}

Diagnostic evidence accumulates through successive interactions with the Medical Visual Tools. The VLM combines observations across views and scales, compares candidate findings, and revisits earlier measurements to verify its conclusions. These operations require access to the original images and measurement results. The finite context window limits the evidence retained in the conversation, while text summaries may omit visual details and numerical information. \ours therefore maintains a lossless memory outside the active context, preserving acquired evidence for subsequent retrieval and assessment.

Each image presented to the VLM is saved with its location and viewing parameters. Each measurement result is stored verbatim with the tool name and arguments used to obtain it. Images and measurements receive unique identifiers. The \texttt{history} tool lists the stored records, and \texttt{recall} retrieves selected images and measurement results by identifier. Multiple images can be retrieved together for cross-view comparison, and previous measurements can be consulted during candidate verification and answer revision.

The memory also retains working notes, findings, and the current answer across context resets. The VLM uses \texttt{note} to record intermediate conclusions, evidence references, and pending checks. Before a reset, it generates a handoff summary of its progress and remaining tasks. The new context includes this summary, working notes, findings, and recent images. Earlier images and measurements remain accessible through \texttt{recall} for subsequent verification and answer revision.

\subsection{Evidence Map}
\label{sec:evidence_map}

Medical images often contain small and sparsely distributed findings across large spatial regions. Examining a region does not necessarily rule out missed findings, as the resolution of an acquired view determines the level of detail available for inspection. A binary record of visited and unvisited regions is therefore insufficient to characterize the progress of image inspection. The Evidence Map addresses this problem by maintaining a spatial representation of the inspected image and estimating the findings that may remain undetected in each region.

Given a target region, the Evidence Map partitions it into spatial cells and assigns each cell an initial estimate of potentially undetected findings based on its size and a prior finding rate. Let $\mathcal{C}$ denote the set of cells, $V_c$ the size of cell $c$, and $\lambda$ the expected finding rate per unit of the image domain. The initial residual estimate is
\begin{equation}
    m_c^{(0)} = \lambda V_c,
    \label{eq:evimap_init}
\end{equation}
where $m_c$ represents the expected number of findings that may remain undetected in cell $c$.

As the VLM examines the image, the Evidence Map updates the residual estimate of each affected cell according to the spatial coverage and resolution of the acquired view. Let $f_{c,t}$ denote the fraction of cell $c$ covered by observation $t$, and $q(r_t)$ the detection weight associated with its viewing resolution $r_t$. The residual estimate is updated as
\begin{equation}
    m_c^{(t+1)}
    =
    m_c^{(t)}\left(1-q(r_t)f_{c,t}\right).
    \label{eq:evimap_update}
\end{equation}
Higher-resolution observations receive larger detection weights, while partial observations reduce the residual estimate only in proportion to the area or volume covered. The total expected number of potentially undetected findings across the target region is
\begin{equation}
    M_{\mathrm{res}} = \sum_{c\in\mathcal{C}} m_c.
    \label{eq:evimap_residual}
\end{equation}

The Evidence Map updates these estimates throughout image inspection without directing the initial search. After the VLM submits its initial answer, the accumulated residual estimate is evaluated against a predefined tolerance. If the estimated number of potentially undetected findings remains above the tolerance, the Evidence Map identifies regions for further inspection.

\subsection{Evidence Verifier}
\label{sec:evidence_verifier}

Visual similarity alone may not be sufficient to distinguish a pathological finding from a normal anatomical structure. For example, a vessel viewed in cross-section may resemble a pulmonary nodule in an axial CT slice. Reliable assessment therefore requires examining suspected findings in greater detail and considering additional visual and quantitative evidence. Moreover, judgments based on different levels of inspection provide different degrees of confidence. The Evidence Verifier addresses these issues by maintaining a belief for each candidate finding and updating it according to the evidence acquired during inspection.

After the initial answer is submitted, the VLM registers the finding addressed by its answer, together with other observed structures that could be confused with it. Each candidate is assigned an identifier and associated with its spatial location and quantitative readouts. The \texttt{register} tool extracts local image characteristics around the candidate, providing measurements for candidate comparison. The VLM then examines each candidate and records its verdict and confidence. This allows the system to assess the reported finding against other plausible findings.

For candidate $j$, let $C_j$ denote the candidate and $p_j^{(t)}$ its current probability of being the target finding. Each candidate is initialized with a prior probability $p_j^{(0)}$, determined by the task and the source of the candidate. The Evidence Verifier updates this probability using the VLM's judgment, with confidence $s_j$ and inspection reliability $\rho_j$ taken into account. The corresponding likelihood ratio is defined as:
\begin{equation}
    \operatorname{LR}(s_j,\rho_j)
    =
    \frac{\rho_j s_j + (1-\rho_j)/2}
         {\rho_j(1-s_j) + (1-\rho_j)/2}.
    \label{eq:evicheck_lr}
\end{equation}
The posterior probability is updated in odds form:
\begin{equation}
    \frac{p_j^{(t+1)}}{1-p_j^{(t+1)}}
    =
    \frac{p_j^{(t)}}{1-p_j^{(t)}}
    \operatorname{LR}(s_j,\rho_j).
    \label{eq:evicheck_update}
\end{equation}
The reliability $\rho_j$ reflects the thoroughness of inspection and determines the weight assigned to each verdict. The Evidence Verifier updates candidate probabilities, allowing findings to be compared and uncertain cases to be identified.

In addition to candidate verification, the Evidence Verifier checks that numerical values reported in the answer are linked to the corresponding measurement tool outputs. Since candidate-registration readouts may include surrounding tissue, reported dimensions and volumes are obtained through separate measurements of the finding itself.

\subsection{Evidence-Guided Agentic Loop}
\label{sec:loop}

In conventional tool-calling loops, VLMs select tools based on their current observations and continue interacting with the image until they produce an answer. However, the model's decision to finish does not necessarily indicate sufficient image inspection or reliable finding identification. \ours therefore introduces an evidence-guided verification stage into the agentic loop, using the Evidence Map and Evidence Verifier to assess the model's initial answer and guide subsequent actions.

The loop begins with autonomous image inspection. The VLM uses the Medical Visual Tools to explore the image and acquire measurements, while the Lossless Evidence Memory preserves the acquired evidence and the Evidence Map continuously updates the status of inspected regions. The initial exploration is not guided by coverage feedback, allowing the VLM to form its own assessment. Once the VLM submits an initial answer, the verification stage begins. The Evidence Verifier evaluates the reported findings and their plausible alternatives, while the Evidence Map assesses the remaining search coverage and identifies regions requiring further inspection.

The resulting feedback is returned to the VLM, which performs additional inspection or measurement and revises its answer as needed. Each subsequent submission triggers another evidence assessment, with the Lossless Evidence Memory keeping previously acquired evidence available throughout the process. The loop terminates once the residual is within the tolerance and no candidate remains undecided, or after at most three verification rounds.

\section{Experiments}
\label{sec:exp}

\subsection{Datasets and Evaluation Metrics}
\label{sec:setup}

\begin{table*}[t]
\centering
\caption{\textbf{Main results by task.} Tasks are grouped into measurement, localization and measurement, detection, and diagnosis. (A) Medical VLMs with direct image input. (B) Methods sharing the Qwen3.8-27B backbone. (C) General VLMs evaluated with and without \ours. Scores are reported in \%; $\Delta$ denotes the change in percentage points relative to direct input with the same backbone.}
\label{tab:main}
\begingroup
\small
\setlength{\tabcolsep}{3.0pt}
\renewcommand{\arraystretch}{1.0}
\setlength{\heavyrulewidth}{0.8pt}
\setlength{\lightrulewidth}{0.4pt}
\setlength{\cmidrulewidth}{0.3pt}
\setlength{\aboverulesep}{1.5pt}
\setlength{\belowrulesep}{1.5pt}
\setlength{\bigstrutjot}{3.3pt}
\definecolor{EviGain}{HTML}{1A7A50}
\definecolor{EviLoss}{HTML}{A64237}
\definecolor{EviNeutral}{HTML}{666666}
\definecolor{EviShade}{gray}{0.94}
\begin{NiceTabular}{l *{11}{c}}
\CodeBefore
  \rowcolor{EviShade}{16,19,22,25,28,31}
\Body
\toprule
\multirow[c]{4}[1]{*}{Model / method}
 & \multicolumn{4}{c}{\textbf{Measurement}}
 & \multicolumn{4}{c}{\textbf{Localization + Measurement}}
 & \textbf{Detection}
 & \multicolumn{2}{c}{\textbf{Diagnosis}} \\
\cmidrule(lr){2-5}\cmidrule(lr){6-9}\cmidrule(lr){10-10}\cmidrule(lr){11-12}
 & \multirow[c]{2}{*}{Phantom}
 & \multirow[c]{2}{*}{Spleen}
 & Lung & Renal & Renal & Lung & Brain
 & \multirow[c]{2}{*}{Echo}
 & \multirow[c]{2}{*}{Nodules}
 & \multirow[c]{2}{*}{CT-RATE}
 & \multirow[c]{2}{*}{CAMELYON16} \\
 & & & tumour & lesion & search & search & tumour & & & & \\
 & {\scriptsize CT} & {\scriptsize CT} & {\scriptsize CT} & {\scriptsize CT} & {\scriptsize CT}& {\scriptsize CT}& {\scriptsize MRI}& {\scriptsize US}& {\scriptsize CT}& {\scriptsize CT}& {\scriptsize Pathology} \\
\midrule
\multicolumn{12}{l}{\textbf{(A)}\enspace\textit{Medical VLMs}} \\
\addlinespace[1pt]
MedGemma-1.5-4B~\cite{sellergren2026medgemma15} & \makebox[2.6em][r]{38.0} & \makebox[2.6em][r]{23.3} & \makebox[2.6em][r]{48.9} & \makebox[2.6em][r]{46.7} & \makebox[2.6em][r]{48.8} & \makebox[2.6em][r]{43.8} & \makebox[2.6em][r]{43.2} & \makebox[2.6em][r]{6.8} & \makebox[2.6em][r]{0.0} & \makebox[2.6em][r]{14.3} & \makebox[2.6em][r]{56.2} \\
MedGemma-27B~\cite{medgemma} & \makebox[2.6em][r]{16.0} & \makebox[2.6em][r]{33.3} & \makebox[2.6em][r]{62.2} & \makebox[2.6em][r]{48.9} & \makebox[2.6em][r]{45.0} & \makebox[2.6em][r]{37.5} & \makebox[2.6em][r]{28.5} & \makebox[2.6em][r]{15.5} & \makebox[2.6em][r]{0.0} & \makebox[2.6em][r]{8.3} & \makebox[2.6em][r]{\textbf{81.2}} \\
Lingshu-32B~\cite{lingshu} & \makebox[2.6em][r]{51.0} & \makebox[2.6em][r]{36.7} & \makebox[2.6em][r]{\textbf{80.0}} & \makebox[2.6em][r]{71.1} & \makebox[2.6em][r]{50.0} & \makebox[2.6em][r]{\underline{50.0}} & \makebox[2.6em][r]{65.0} & \makebox[2.6em][r]{27.5} & \makebox[2.6em][r]{0.0} & \makebox[2.6em][r]{13.1} & \makebox[2.6em][r]{50.0} \\
Hulu-Med-32B~\cite{hulumed} & \makebox[2.6em][r]{36.0} & \makebox[2.6em][r]{31.7} & \makebox[2.6em][r]{66.7} & \makebox[2.6em][r]{53.3} & \makebox[2.6em][r]{51.2} & \makebox[2.6em][r]{33.8} & \makebox[2.6em][r]{33.2} & \makebox[2.6em][r]{6.0} & \makebox[2.6em][r]{0.0} & \makebox[2.6em][r]{\textbf{48.0}} & \makebox[2.6em][r]{50.0} \\
ClinFusion-32B~\cite{clinfusion} & \makebox[2.6em][r]{42.0} & \makebox[2.6em][r]{46.7} & \makebox[2.6em][r]{60.0} & \makebox[2.6em][r]{71.1} & \makebox[2.6em][r]{51.2} & \makebox[2.6em][r]{\textbf{68.8}} & \makebox[2.6em][r]{55.2} & \makebox[2.6em][r]{20.5} & \makebox[2.6em][r]{0.0} & \makebox[2.6em][r]{\underline{34.6}} & \makebox[2.6em][r]{53.1} \\
\midrule
\multicolumn{12}{l}{\textbf{(B)}\enspace\textit{Qwen3.8-27B Backbone}} \\
\addlinespace[1pt]
Direct input & \makebox[2.6em][r]{77.0} & \makebox[2.6em][r]{63.3} & \makebox[2.6em][r]{64.4} & \makebox[2.6em][r]{71.1} & \makebox[2.6em][r]{57.5} & \makebox[2.6em][r]{37.5} & \makebox[2.6em][r]{68.2} & \makebox[2.6em][r]{37.2} & \makebox[2.6em][r]{0.0} & \makebox[2.6em][r]{24.5} & \makebox[2.6em][r]{56.2} \\
Zoom agent & \makebox[2.6em][r]{\underline{97.0}} & \makebox[2.6em][r]{71.7} & \makebox[2.6em][r]{\underline{75.6}} & \makebox[2.6em][r]{91.1} & \makebox[2.6em][r]{53.8} & \makebox[2.6em][r]{28.7} & \makebox[2.6em][r]{75.0} & \makebox[2.6em][r]{37.5} & \makebox[2.6em][r]{\underline{14.1}} & \makebox[2.6em][r]{26.7} & \makebox[2.6em][r]{53.1} \\
VISTA~\cite{vista} & \makebox[2.6em][r]{94.0} & \makebox[2.6em][r]{\underline{90.0}} & \makebox[2.6em][r]{71.1} & \makebox[2.6em][r]{\underline{93.3}} & \makebox[2.6em][r]{58.8} & \makebox[2.6em][r]{27.5} & \makebox[2.6em][r]{\underline{79.2}} & \makebox[2.6em][r]{32.2} & \makebox[2.6em][r]{9.4} & \makebox[2.6em][r]{25.9} & \makebox[2.6em][r]{53.1} \\
MDAgents~\cite{mdagents} & \makebox[2.6em][r]{81.0} & \makebox[2.6em][r]{63.3} & \makebox[2.6em][r]{66.7} & \makebox[2.6em][r]{71.1} & \makebox[2.6em][r]{\underline{60.0}} & \makebox[2.6em][r]{47.5} & \makebox[2.6em][r]{63.2} & \makebox[2.6em][r]{\underline{37.8}} & \makebox[2.6em][r]{0.0} & \makebox[2.6em][r]{31.7} & \makebox[2.6em][r]{\underline{65.6}} \\
\textbf{\ours} & \makebox[2.6em][r]{\textbf{100.0}} & \makebox[2.6em][r]{\textbf{98.3}} & \makebox[2.6em][r]{\underline{75.6}} & \makebox[2.6em][r]{\textbf{96.7}} & \makebox[2.6em][r]{\textbf{71.2}} & \makebox[2.6em][r]{40.0} & \makebox[2.6em][r]{\textbf{85.8}} & \makebox[2.6em][r]{\textbf{46.2}} & \makebox[2.6em][r]{\textbf{28.2}} & \makebox[2.6em][r]{33.6} & \makebox[2.6em][r]{\underline{65.6}} \\
\specialrule{\lightrulewidth}{0pt}{\belowrulesep} 
\multicolumn{12}{l}{\textbf{(C)}\enspace\textit{Cross-backbone evaluation}} \\
\addlinespace[1pt]
Gemma 4-31B & \makebox[2.6em][r]{73.0} & \makebox[2.6em][r]{53.3} & \makebox[2.6em][r]{62.2} & \makebox[2.6em][r]{71.1} & \makebox[2.6em][r]{58.8} & \makebox[2.6em][r]{58.8} & \makebox[2.6em][r]{64.8} & \makebox[2.6em][r]{39.8} & \makebox[2.6em][r]{3.5} & \makebox[2.6em][r]{30.5} & \makebox[2.6em][r]{43.8} \\
\hspace{0.7em}\textbf{+ \ours} & \makebox[2.6em][r]{84.0} & \makebox[2.6em][r]{41.7} & \makebox[2.6em][r]{64.4} & \makebox[2.6em][r]{76.7} & \makebox[2.6em][r]{53.8} & \makebox[2.6em][r]{45.0} & \makebox[2.6em][r]{53.2} & \makebox[2.6em][r]{33.5} & \makebox[2.6em][r]{29.4} & \makebox[2.6em][r]{29.6} & \makebox[2.6em][r]{59.4} \\
\hspace{0.7em}$\Delta$ (pp) & \makebox[2.6em][r]{\textcolor{EviGain}{+11.0}} & \makebox[2.6em][r]{\textcolor{EviLoss}{\textminus{}11.6}} & \makebox[2.6em][r]{\textcolor{EviGain}{+2.2}} & \makebox[2.6em][r]{\textcolor{EviGain}{+5.6}} & \makebox[2.6em][r]{\textcolor{EviLoss}{\textminus{}5.0}} & \makebox[2.6em][r]{\textcolor{EviLoss}{\textminus{}13.8}} & \makebox[2.6em][r]{\textcolor{EviLoss}{\textminus{}11.6}} & \makebox[2.6em][r]{\textcolor{EviLoss}{\textminus{}6.3}} & \makebox[2.6em][r]{\textcolor{EviGain}{+25.9}} & \makebox[2.6em][r]{\textcolor{EviLoss}{\textminus{}0.9}} & \makebox[2.6em][r]{\textcolor{EviGain}{+15.6}} \\
\addlinespace[2.5pt]

Gemini 3.5 Flash$^\dagger$ & \makebox[2.6em][r]{81.0} & \makebox[2.6em][r]{73.3} & \makebox[2.6em][r]{62.2} & \makebox[2.6em][r]{83.3} & \makebox[2.6em][r]{52.5} & \makebox[2.6em][r]{52.5} & \makebox[2.6em][r]{77.9} & \makebox[2.6em][r]{26.2} & \makebox[2.6em][r]{0.0} & \makebox[2.6em][r]{34.3} & \makebox[2.6em][r]{50.0} \\
\hspace{0.7em}\textbf{+ \ours} & \makebox[2.6em][r]{100.0} & \makebox[2.6em][r]{96.7} & \makebox[2.6em][r]{86.7} & \makebox[2.6em][r]{94.4} & \makebox[2.6em][r]{67.5} & \makebox[2.6em][r]{78.8} & \makebox[2.6em][r]{84.2} & \makebox[2.6em][r]{45.8} & \makebox[2.6em][r]{58.5} & \makebox[2.6em][r]{38.9} & \makebox[2.6em][r]{62.5} \\
\hspace{0.7em}$\Delta$ (pp) & \makebox[2.6em][r]{\textcolor{EviGain}{+19.0}} & \makebox[2.6em][r]{\textcolor{EviGain}{+23.4}} & \makebox[2.6em][r]{\textcolor{EviGain}{+24.5}} & \makebox[2.6em][r]{\textcolor{EviGain}{+11.1}} & \makebox[2.6em][r]{\textcolor{EviGain}{+15.0}} & \makebox[2.6em][r]{\textcolor{EviGain}{+26.3}} & \makebox[2.6em][r]{\textcolor{EviGain}{+6.3}} & \makebox[2.6em][r]{\textcolor{EviGain}{+19.6}} & \makebox[2.6em][r]{\textcolor{EviGain}{+58.5}} & \makebox[2.6em][r]{\textcolor{EviGain}{+4.6}} & \makebox[2.6em][r]{\textcolor{EviGain}{+12.5}} \\
\addlinespace[2.5pt]

Kimi K3$^\dagger$ & \makebox[2.6em][r]{77.0} & \makebox[2.6em][r]{83.3} & \makebox[2.6em][r]{75.6} & \makebox[2.6em][r]{67.8} & \makebox[2.6em][r]{57.5} & \makebox[2.6em][r]{42.5} & \makebox[2.6em][r]{72.5} & \makebox[2.6em][r]{29.6} & \makebox[2.6em][r]{0.0} & \makebox[2.6em][r]{22.2} & \makebox[2.6em][r]{62.5} \\
\hspace{0.7em}\textbf{+ \ours} & \makebox[2.6em][r]{99.0} & \makebox[2.6em][r]{96.7} & \makebox[2.6em][r]{88.9} & \makebox[2.6em][r]{95.6} & \makebox[2.6em][r]{62.5} & \makebox[2.6em][r]{70.0} & \makebox[2.6em][r]{87.5} & \makebox[2.6em][r]{55.4} & \makebox[2.6em][r]{27.7} & \makebox[2.6em][r]{37.0} & \makebox[2.6em][r]{62.5} \\
\hspace{0.7em}$\Delta$ (pp) & \makebox[2.6em][r]{\textcolor{EviGain}{+22.0}} & \makebox[2.6em][r]{\textcolor{EviGain}{+13.4}} & \makebox[2.6em][r]{\textcolor{EviGain}{+13.3}} & \makebox[2.6em][r]{\textcolor{EviGain}{+27.8}} & \makebox[2.6em][r]{\textcolor{EviGain}{+5.0}} & \makebox[2.6em][r]{\textcolor{EviGain}{+27.5}} & \makebox[2.6em][r]{\textcolor{EviGain}{+15.0}} & \makebox[2.6em][r]{\textcolor{EviGain}{+25.8}} & \makebox[2.6em][r]{\textcolor{EviGain}{+27.7}} & \makebox[2.6em][r]{\textcolor{EviGain}{+14.8}} & \makebox[2.6em][r]{\textcolor{EviNeutral}{+0.0}} \\
\addlinespace[2.5pt]

Claude Opus 5.5$^\dagger$ & \makebox[2.6em][r]{87.0} & \makebox[2.6em][r]{88.3} & \makebox[2.6em][r]{73.3} & \makebox[2.6em][r]{73.3} & \makebox[2.6em][r]{61.3} & \makebox[2.6em][r]{55.0} & \makebox[2.6em][r]{87.9} & \makebox[2.6em][r]{56.2} & \makebox[2.6em][r]{6.2} & \makebox[2.6em][r]{35.1} & \makebox[2.6em][r]{71.9} \\
\hspace{0.7em}\textbf{+ \ours} & \makebox[2.6em][r]{100.0} & \makebox[2.6em][r]{100.0} & \makebox[2.6em][r]{91.1} & \makebox[2.6em][r]{97.8} & \makebox[2.6em][r]{81.2} & \makebox[2.6em][r]{78.8} & \makebox[2.6em][r]{91.2} & \makebox[2.6em][r]{65.4} & \makebox[2.6em][r]{60.0} & \makebox[2.6em][r]{38.6} & \makebox[2.6em][r]{81.2} \\
\hspace{0.7em}$\Delta$ (pp) & \makebox[2.6em][r]{\textcolor{EviGain}{+13.0}} & \makebox[2.6em][r]{\textcolor{EviGain}{+11.7}} & \makebox[2.6em][r]{\textcolor{EviGain}{+17.8}} & \makebox[2.6em][r]{\textcolor{EviGain}{+24.5}} & \makebox[2.6em][r]{\textcolor{EviGain}{+19.9}} & \makebox[2.6em][r]{\textcolor{EviGain}{+23.8}} & \makebox[2.6em][r]{\textcolor{EviGain}{+3.3}} & \makebox[2.6em][r]{\textcolor{EviGain}{+9.2}} & \makebox[2.6em][r]{\textcolor{EviGain}{+53.8}} & \makebox[2.6em][r]{\textcolor{EviGain}{+3.5}} & \makebox[2.6em][r]{\textcolor{EviGain}{+9.3}} \\
\addlinespace[2.5pt]

GPT-6.1 Sol$^\dagger$ & \makebox[2.6em][r]{93.0} & \makebox[2.6em][r]{98.3} & \makebox[2.6em][r]{77.8} & \makebox[2.6em][r]{80.0} & \makebox[2.6em][r]{66.2} & \makebox[2.6em][r]{63.7} & \makebox[2.6em][r]{86.7} & \makebox[2.6em][r]{70.8} & \makebox[2.6em][r]{10.8} & \makebox[2.6em][r]{41.4} & \makebox[2.6em][r]{53.1} \\
\hspace{0.7em}\textbf{+ \ours} & \makebox[2.6em][r]{100.0} & \makebox[2.6em][r]{100.0} & \makebox[2.6em][r]{95.6} & \makebox[2.6em][r]{97.8} & \makebox[2.6em][r]{87.5} & \makebox[2.6em][r]{81.2} & \makebox[2.6em][r]{95.4} & \makebox[2.6em][r]{63.7} & \makebox[2.6em][r]{69.2} & \makebox[2.6em][r]{49.2} & \makebox[2.6em][r]{78.1} \\
\hspace{0.7em}$\Delta$ (pp) & \makebox[2.6em][r]{\textcolor{EviGain}{+7.0}} & \makebox[2.6em][r]{\textcolor{EviGain}{+1.7}} & \makebox[2.6em][r]{\textcolor{EviGain}{+17.8}} & \makebox[2.6em][r]{\textcolor{EviGain}{+17.8}} & \makebox[2.6em][r]{\textcolor{EviGain}{+21.3}} & \makebox[2.6em][r]{\textcolor{EviGain}{+17.5}} & \makebox[2.6em][r]{\textcolor{EviGain}{+8.7}} & \makebox[2.6em][r]{\textcolor{EviLoss}{\textminus{}7.1}} & \makebox[2.6em][r]{\textcolor{EviGain}{+58.4}} & \makebox[2.6em][r]{\textcolor{EviGain}{+7.8}} & \makebox[2.6em][r]{\textcolor{EviGain}{+25.0}} \\

\bottomrule
\end{NiceTabular}
\par\vspace{3pt}
\begin{minipage}{\textwidth}
\footnotesize
\textbf{Bold} and \underline{underline}: best and second-best scores across (A) and (B). Shaded rows use \ours. Positive and negative $\Delta$ values denote gains and drops relative to direct input with the same backbone. $^\dagger$ API-based models were evaluated on reduced subsets
for LUNA16, brain tumour MRI, and echocardiography.
\end{minipage}
\endgroup
\end{table*}

\noindent\textbf{Datasets.}
We evaluate \ours on 11 tasks spanning CT, MRI, ultrasound, and whole-slide pathology. The tasks are grouped into four categories: measurement, localization and measurement, detection, and diagnosis.

\emph{Measurement.} Four tasks evaluate quantitative measurements with the target structure or its location specified in the prompt. These include synthetic phantoms (20 cases), spleen and lung tumour CT from MSD~\cite{msd} (15 cases each), and renal lesion CT from KiTS23~\cite{kits} (30 cases). The requested measurements include diameter, volume, and Hounsfield units. The renal lesion task also includes differentiation between cysts and solid tumours.

\emph{Localization and measurement.} Four tasks require the VLM to search for and localize the target before performing quantitative measurement. These include renal search on KiTS23~\cite{kits} (20 cases), lung search on MSD~\cite{msd} (20 cases), brain tumour MRI from BraTS/MSD~\cite{msd,brats} (100 cases), and echocardiography from CAMUS~\cite{camus} (100 cases). Successful completion depends on both correct target localization and accurate measurement.

\emph{Detection.} The detection task targets all pulmonary nodules with a diameter of at least 3\,mm in 69 held-out CT scans from LUNA16~\cite{luna16}. The reference set contains 85 nodules accepted by at least three of four readers.

\emph{Diagnosis.} The two diagnosis tasks are multi-label classification of 18 chest CT findings on CT-RATE~\cite{ctrate} (50 scans) and slide-level classification of sentinel lymph node metastasis on CAMELYON16~\cite{camelyon16} (24 whole-slide images).

\noindent\textbf{Evaluation metrics.}
For measurement and localization-and-measurement tasks, a numerical prediction is counted as correct if its error falls within the predefined clinical tolerance for that quantity. Categorical predictions, such as cyst versus solid, are evaluated by exact agreement with the reference label. The score for each case is the proportion of correct predictions across the requested outputs. For LUNA16~\cite{luna16}, we report nodule-level sensitivity for detections with confidence $\geq 0.5$. A prediction is matched to a reference nodule if the distance between their centers is within the reference nodule's radius. For CT-RATE, we report macro-F1 across the 18 chest findings. For CAMELYON16, we report balanced accuracy for slide-level metastasis classification.

\begin{figure*}[ht]
\centering
\includegraphics[width=\textwidth]{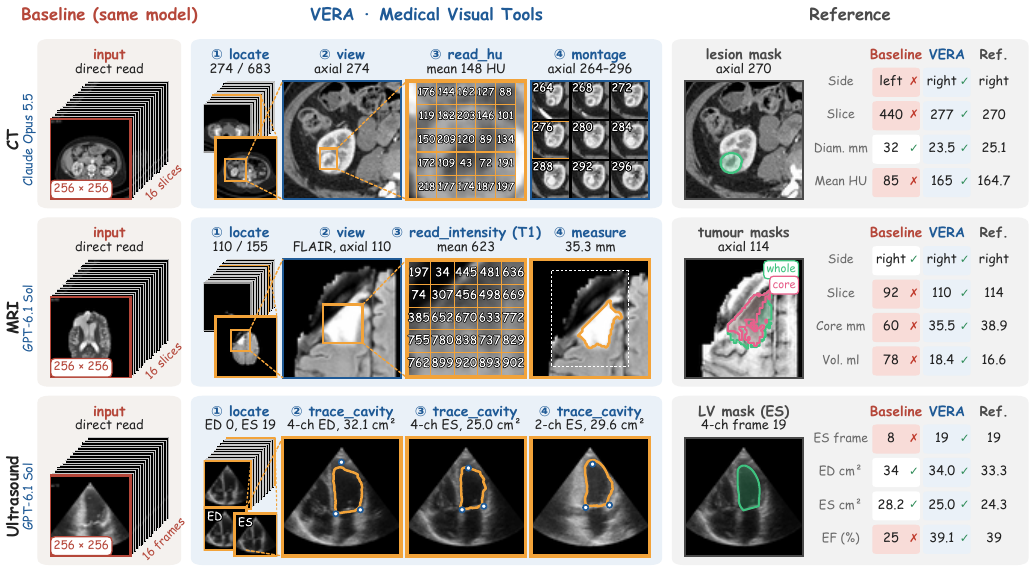}
\caption{\textbf{Qualitative comparison on CT, MRI and ultrasound.} Each row compares a general VLM reading images directly (left: its 16-slice or 16-frame input) with that VLM in \ours (middle: it locates the slice or frames in the full sequence, zooms in, and measures; its tool calls and readouts), against the reference annotation (right). Wrong fields are red.
CT (Claude Opus 5.5): largest solid renal tumour (side, slice, diameter, mean HU).
MRI (GPT-6.1 Sol): glioma (side, slice of largest core, core diameter, whole-tumour volume).
Ultrasound (GPT-6.1 Sol): end-systolic frame, LV areas and ejection fraction.}
\label{fig:qualitative}
\end{figure*}

\subsection{Implementation Details}
\label{sec:impl}

\noindent\textbf{Backbones.}
We use Qwen3.8-27B-FP8 as the primary backbone, deployed locally with vLLM. Additional backbones include the open-source VLM Gemma~4 31B and four API-based models, Gemini 3.5 Flash, Kimi K3, Claude Opus 5.5, and GPT-6.1 Sol. Each backbone is evaluated with direct image input and with \ours on the same cases (\Cref{tab:main}, bottom). Due to the API costs for three token-intensive tasks, we evaluate the four API-based models on smaller subsets: 40 LUNA16 scans, 60 brain tumour MRI cases, and 60 echocardiography cases.

\noindent\textbf{Baselines.}
We compare \ours with two groups of baselines using the same cases, prompts, and step budgets.
\emph{Medical VLMs.} We evaluate five medical VLMs with direct image input and no agent framework: MedGemma-1.5-4B~\cite{sellergren2026medgemma15} and MedGemma-27B~\cite{medgemma}, Lingshu-32B~\cite{lingshu}, Hulu-Med-32B~\cite{hulumed}, and ClinFusion-32B~\cite{clinfusion}. Each model is evaluated using its supported input formats, including native 3D loading, labeled 2D slices, or frames. For models supporting multiple formats, we report the best result across those formats.
\emph{Agent baselines.} All three baselines use the same Qwen3.8-27B backbone as \ours. The zoom agent follows ``thinking with images''~\cite{wu2024vstar,deepeyes}, supporting slice rendering, panning, and cropping without structured memory. VISTA~\cite{vista} combines lossless visual memory with tools for retrieval, zooming, and quantitative readouts. MDAgents~\cite{mdagents} is a multi-agent framework that rates the complexity of each case and assigns either a single clinician agent or a team of clinician agents, all reading the same images as the direct read.

\subsection{Main Results}
\label{sec:main}

\begin{table*}[ht]
\centering
\caption{\textbf{Ablation results on the Qwen3.8-27B backbone across four task categories.} Small numbers indicate changes relative to the full method.}
\label{tab:abl}
\begingroup
\small
\setlength{\tabcolsep}{1.85pt}
\renewcommand{\arraystretch}{1.04}
\setlength{\heavyrulewidth}{0.8pt}
\setlength{\lightrulewidth}{0.4pt}
\setlength{\cmidrulewidth}{0.3pt}
\setlength{\aboverulesep}{1.5pt}
\setlength{\belowrulesep}{1.5pt}
\definecolor{EviShade}{gray}{0.94}
\begin{NiceTabular}{l *{11}{c}}
\CodeBefore
  \rowcolor{EviShade}{4}
\Body
\toprule
\multirow[c]{3}{*}{Variant}
 & \multicolumn{4}{c}{\textbf{Measurement}}
 & \multicolumn{4}{c}{\textbf{Loc.\ + Measurement}}
 & \textbf{Det.}
 & \multicolumn{2}{c}{\textbf{Diagnosis}} \\
\cmidrule(lr){2-5}\cmidrule(lr){6-9}\cmidrule(lr){10-10}\cmidrule(lr){11-12}
 & \multirow[c]{2}{*}{Phantom}
 & \multirow[c]{2}{*}{Spleen}
 & Lung
 & Renal
 & Renal
 & Lung
 & Brain
 & \multirow[c]{2}{*}{Echo}
 & \multirow[c]{2}{*}{Nodules}
 & \multirow[c]{2}{*}{CT-RATE}
 & \multirow[c]{2}{*}{CAMELYON16} \\
 & &
 & tumour
 & lesion
 & search
 & search
 & tumour
 & &
 & &\\
\midrule
\ours (full) & 100.0 & 98.3 & 75.6 & 96.7 & 71.2 & 40.0 & 85.8 & 46.2 & 28.2 & 33.6 & 65.6 \\
w/o Evidence Verifier & 96.0\,{\tiny $-$4.0} & 70.0\,{\tiny $-$28.3} & 82.2\,{\tiny +6.6} & 85.6\,{\tiny $-$11.1} & 60.0\,{\tiny $-$11.2} & 28.7\,{\tiny $-$11.3} & 78.8\,{\tiny $-$7.0} & 39.0\,{\tiny $-$7.2} & 25.9\,{\tiny $-$2.3} & 38.4\,{\tiny +4.8} & 53.1\,{\tiny $-$12.5} \\
w/o Evidence Map & 97.0\,{\tiny $-$3.0} & 95.0\,{\tiny $-$3.3} & 75.6\,{\tiny 0.0} & 94.4\,{\tiny $-$2.3} & 65.0\,{\tiny $-$6.2} & 31.2\,{\tiny $-$8.8} & 83.2\,{\tiny $-$2.6} & 43.2\,{\tiny $-$3.0} & 14.1\,{\tiny $-$14.1} & 30.0\,{\tiny $-$3.6} & 59.4\,{\tiny $-$6.2} \\
w/o both & 97.0\,{\tiny $-$3.0} & 70.0\,{\tiny $-$28.3} & 73.3\,{\tiny $-$2.3} & 90.0\,{\tiny $-$6.7} & 55.0\,{\tiny $-$16.2} & 30.0\,{\tiny $-$10.0} & 77.0\,{\tiny $-$8.8} & 33.5\,{\tiny $-$12.7} & 12.9\,{\tiny $-$15.3} & 30.0\,{\tiny $-$3.6} & 56.2\,{\tiny $-$9.4} \\
\bottomrule
\end{NiceTabular}
\endgroup
\end{table*}

\noindent\textbf{Comparison with medical VLMs.}
The five medical VLMs show varying performance across tasks. Lingshu-32B achieves the highest score on lung tumour measurement (80.0\%), while ClinFusion-32B performs best on lung search (68.8\%). MedGemma-27B leads on CAMELYON16 (81.2\%), and Hulu-Med-32B achieves the highest CT-RATE score (48.0\%). However, none of the five models detects pulmonary nodules, and their echocardiography scores do not exceed 27.5\%. In comparison, VERA with the Qwen3.8-27B backbone surpasses all five medical VLMs on 7 of the 11 tasks. The gains are particularly clear in tasks requiring localization and measurement: VERA achieves 71.2\% on renal search and 85.8\% on brain tumour measurement, compared with the best medical VLM scores of 51.2\% and 65.0\%, respectively. VERA also achieves 28.2\% on pulmonary nodule detection and 46.2\% on echocardiography, where all five medical VLMs perform poorly.

\noindent\textbf{Comparison with agent baselines.} Zoom agent and VISTA improve measurement through interactive visual inspection: VISTA raises spleen measurement from 63.3\% to 90.0\% and renal lesion measurement from 71.1\% to 93.3\%. Their localization gains are less consistent, and pulmonary nodule detection sensitivity does not exceed 14.1\%. MDAgents improves diagnostic performance but provides limited measurement gains and detects no pulmonary nodules. \ours reaches 28.2\% pulmonary nodule sensitivity, double the best agent baseline, and 71.2\% on renal search versus 58.8\% for VISTA and 60.0\% for MDAgents. It also achieves the highest CT-RATE score (33.6\%) among agent baselines and matches their best CAMELYON16 score (65.6\%), supporting diagnostic evidence assessment within visual interaction.

\noindent\textbf{Performance with stronger backbones.} \ours improves performance across backbones, with particularly strong results on high-capability API-based VLMs, which frequently exceed 90\% on measurement and localization tasks. GPT-6.1 Sol reaches 95.6\% on lung tumour measurement, 87.5\% on renal search, and 95.4\% on brain tumour measurement, substantially outperforming Qwen3.8-27B+VERA on these tasks. Pulmonary nodule detection sensitivity increases from 6.2\% to 60.0\% with Claude Opus 5.5 and from 10.8\% to 69.2\% with GPT-6.1 Sol. These gains show that \ours enables more capable backbones to achieve substantially higher performance across medical imaging tasks. Gemma~4-31B shows less consistent gains.

\subsection{Qualitative Results}
\label{sec:qualitative}

\Cref{fig:qualitative} presents three examples from CT, MRI, and ultrasound. In the CT case, direct image input leads to incorrect lesion localization and inaccurate size and density estimates. With VERA, the VLM locates the lesion, examines the corresponding slice and neighboring slices, and obtains HU measurements from the lesion region, producing results closer to the reference. In the MRI case, VERA locates the tumour and combines T1
intensity readouts with targeted FLAIR measurements to estimate
the core diameter and whole-tumour volume. For echocardiography, VERA identifies the correct end-systolic frame and measures the left ventricle across two views, yielding an ejection fraction of 39.1\%, close to the reference value of 39.0\%. Across these examples, VERA enables the VLM to move beyond fixed image input and obtain task-specific evidence through targeted visual inspection and measurement. Together, these cases illustrate how VERA turns visual interaction into task-directed evidence acquisition, enabling more accurate localization, measurement, and assessment across different imaging modalities.

\subsection{Ablation Studies}
\label{sec:ablation}

We examine the individual and joint contributions of the Evidence Map and Evidence Verifier on the Qwen3.8-27B backbone in Table~\ref{tab:abl}. Removing the Evidence Verifier leads to substantial drops in quantitative measurement, including spleen (98.3\% to 70.0\%) and renal lesion (96.7\% to 85.6\%). Removing the Evidence Map has its largest impact on pulmonary nodule detection, where sensitivity drops from 28.2\% to 14.1\%, as well as lung search (40.0\% to 31.2\%). These results show that the two components address different aspects of medical image interpretation: assessing identified findings and tracking the progress of image inspection. Removing both components also disables Evidence-Guided Agentic Loop, leaving the VLM to rely on its initial assessment without feedback on the acquired evidence. Performance drops across all 11 tasks, with particularly large declines in spleen measurement ($-28.3$ points), renal search ($-16.2$ points), and pulmonary nodule detection ($-15.3$ points). This demonstrates that the Evidence Map and Evidence Verifier work together to guide the agentic loop, connecting visual exploration with evidence assessment and subsequent inspection.

\subsection{Limitations}
\label{sec:limitations}

Our evaluation has several limitations. First, the four API-based VLMs were evaluated on smaller subsets for three token-intensive tasks due to API costs, limiting the scope of cross-backbone comparisons. Second, since our evaluation relies on public datasets, we cannot rule out potential overlap between these datasets and the training data of proprietary VLMs. Third, VERA's performance remains dependent on the VLM's ability to interpret medical images and execute tool calls reliably. Limitations in these capabilities may adversely affect VERA's performance across medical imaging tasks. Finally, our evaluation is based on public datasets and does not include validation in real-world clinical environments. 

\section{Conclusion}
\label{sec:conclusion}

We introduced VERA, a medical visual harness that equips general VLMs with interactive visual tools and systematic diagnostic evidence management. By preserving acquired observations and assessing the evidence supporting identified findings, VERA guides VLMs through iterative visual exploration, verification, and targeted re-examination. Experiments across 11 tasks and 4 imaging modalities demonstrate consistent improvements over direct image input, spanning measurement, localization, detection, and diagnosis. These improvements are also observed with strong API-based VLMs, which achieve substantially higher performance when equipped with VERA. Our findings highlight the potential of visual harnesses as a new direction for advancing general VLMs in medical imaging.

\clearpage



{
    \small
    \bibliographystyle{ieeenat_fullname}
    \bibliography{main}
}

\end{document}